\documentclass[10pt,twocolumn,letterpaper]{article}

\usepackage[pagenumbers]{cvpr}            
\usepackage{wrapfig}
\definecolor{green_zjr}{RGB}{6,104,86}
\usepackage[dvipsnames]{xcolor}
\usepackage{pifont}  

\newcommand{\et}[2]{${#1}^{\pm{#2}}$}
\newcommand{\etb}[2]{$\mathbf{{#1}}^{\pm{#2}}$}

\definecolor{LightGray}{rgb}{0.9,0.9,0.9}
\definecolor{DarkRed}{rgb}{0.75,0,0}
\definecolor{DarkBlue}{rgb}{0,0,0.55}
\definecolor{DarkGreen}{rgb}{0.43, 0.68, 0.28}

\usepackage{multirow}

\usepackage{amsmath,amsfonts,bm}

\def\eqref#1{equation~\ref{#1}}

\def\1{\bm{1}}

\def\vc{{\bm{c}}}

\def\vx{{\bm{x}}}

\def\mK{{\bm{K}}}

\def\mQ{{\bm{Q}}}

\def\mV{{\bm{V}}}

\def\mX{{\bm{X}}}

\DeclareMathAlphabet{\mathsfit}{\encodingdefault}{\sfdefault}{m}{sl}
\SetMathAlphabet{\mathsfit}{bold}{\encodingdefault}{\sfdefault}{bx}{n}

\def\sC{{\mathbb{C}}}

\def\sZ{{\mathbb{Z}}}

\newcommand{\softmax}{\mathrm{softmax}}

\definecolor{cvprblue}{rgb}{0.21,0.49,0.74}
\usepackage[pagebackref,breaklinks,colorlinks,citecolor=cvprblue]{hyperref}

\def\paperID{***} 
\def\confName{CVPR}
\def\confYear{2025}

\title{\textsc{EnergyMoGen}: Compositional Human Motion Generation  with \\ Energy-Based  Diffusion Model in Latent Space}

\author{Jianrong Zhang$^{1}$, 
Hehe Fan$^{2,\dagger}$, 
Yi Yang$^2$ \\
\small \qquad $^{\dagger}$Corresponding author \\
\small $^1$ReLER, AAII, University of Technology Sydney \qquad
$^2$CCAI, Zhejiang University \\
\small \url{https://jiro-zhang.github.io/EnergyMoGen/}
}

\begin{document}
\maketitle
\vspace{-1.5mm}
\begin{abstract}

Diffusion models, particularly latent diffusion models, have demonstrated remarkable success in text-driven human motion generation. However, it remains challenging for latent diffusion models to effectively compose multiple semantic concepts into a single, coherent motion sequence. 
To address this issue, we propose EnergyMoGen, which includes two spectrums of Energy-Based Models: \ding{182} We interpret the diffusion model as a latent-aware energy-based model that generates motions by composing a set of diffusion models in latent space; \ding{183} We introduce a semantic-aware energy model based on cross-attention, which enables semantic composition and adaptive gradient descent for text embeddings.
To overcome the challenges of semantic inconsistency and motion distortion across these two spectrums, we introduce Synergistic Energy Fusion. This design allows the motion latent diffusion model to synthesize high-quality, complex motions by combining multiple energy terms corresponding to textual descriptions. 
Experiments show that our approach outperforms existing state-of-the-art models on various motion generation tasks, including text-to-motion generation, compositional motion generation, and multi-concept motion generation. Additionally, we demonstrate that our method can be used to extend motion datasets and improve the text-to-motion task. 

\end{abstract}    
\vspace{-1mm}
\section{Introduction}
\label{sec:intro}

\begin{quote}
\it \small
Composition of complex ideas out of simple ones. \\
\mbox{}\hfill -- John Locke (1632 - 1704)~\textnormal{\cite{1991Locke}}
\vspace{-4pt}
\end{quote}
Generating human motions that are both visually appealing and physically plausible is a longstanding goal in computer animation. Recently, the remarkable success of diffusion models~\cite{sohl2015deep,ho2020denoising,rombach2022high} stimulates the emergence of diffusion-based solutions for generating human motion~\cite{wang2023fg,tevet2022MDM,zhang2023remodiffuse,zhang2023finemogen} from a single text prompt.

Humans are masters of motion composition. For example, with the mind of ``raising both arms'' and ``walking forward'', we can blend these simple concepts to perform complex motions simultaneously based on our experiences. To achieve this capability, past efforts, such as PriorMDM~\cite{shafir2024priormdm} and GMD~\cite{karunratanakul2023guided}, have made significant strides in temporal~\cite{TEACH:3DV:2022} and spatial~\cite{SINC:ICCV:2023} human motion compositions using pre-trained skeleton-based diffusion models. These models operate directly on skeletons, thus enabling precise control over joint feature distributions to generate realistic motions from multiple textual descriptions. 

\begin{figure*}[t]
\centering
\includegraphics[width=\textwidth]{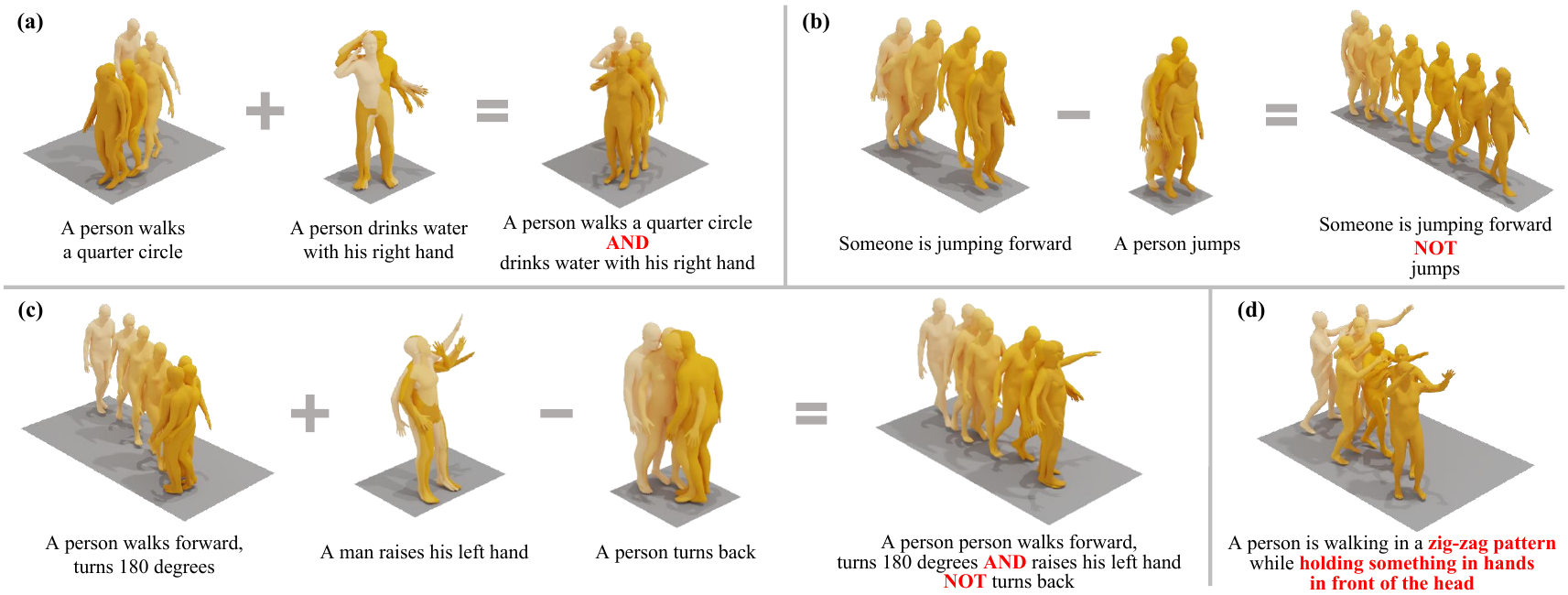}
\vspace{-7.5mm}
\caption{\textbf{Compositional motion generation.} Our approach is able to compose complex motions from simple concepts in settings of (a) concept conjunction, (b) concept negation, (c) compositional motion generation with conjunction and negation, and (d) multi-concept motion generation (a form of compositional generation).
}
\vspace{-4.5mm}
\label{fig:teaser}
\end{figure*} 

In this work, we are interested in the motion latent diffusion model~\cite{rombach2022high,chen2022mld}, which is another important formulation of diffusion models operating in continuous latent space. However, such models are not feasible for composing motion through algorithms proposed in skeleton-based diffusion models. As latent diffusion models usually use one (or a fixed number of) latent vector(s) to represent variable-length motions (\eg, ranging from 40 to 196 frames), the following issues naturally arise: \textit{\textbf{i)}} Lack of explicit correspondence (\eg, one-to-one or one-to-many mapping) between latent vectors and motions; \textit{\textbf{ii)}} It is challenging for latent vectors to support per-frame composition.

Tackling these two issues can offer valuable insights into compositional motion generation algorithms and motivate us to reconsider this task from an energy view. The idea of ``energy'' is intrinsic to Energy-Based Models (EBMs)~\cite{lecun2006tutorial}. Research on EBMs has a long history, yet it gets overlooked nowadays in human motion generation. The energy function defines a potential field to shape the energy surface by assigning low energies to desired configurations and high energies to other configurations~\cite{zhao2017energybased}. With this perspective~\cite{du2019implicit,du2020compositional,du2023reduce,du2020improved,park2024energy}, we delve into two spectrums of EBMs for compositional motion generation: \ding{182} \textbf{\textit{latent-aware}} and \ding{183} \textbf{\textit{semantic-aware}}. By formulating the generative process of latent diffusion models as an energy combination problem, we can compose complex motions from simple components through two compositional operations: \textbf{conjunction} and \textbf{negation} (Figure~\ref{fig:teaser} (a), (b), (c), more visual results are provided in $\S$ \ref{sec:exp_composition} and on the \href{https://jiro-zhang.github.io/EnergyMoGen/}{project page}). 

Concretely, we propose \textsc{EnergyMoGen}, a compositional motion generation framework that incorporates cross-attention within the denoising autoencoder. We view latent diffusion models as latent-aware EBMs and illustrate the connection between them. Leveraging classifier-free guidance~\cite{ho2022classifier}, we are able to compose motions by implicitly combining energy terms  (\ie, latent distributions) of multiple diffusion models. 
In addition, we introduce a semantic-aware EBM based on the cross-attention mechanism that supports a semantic-aware energy mixture for motion composition. Meanwhile, this energy-based cross-attention enables the adaptive updating of text embeddings by calculating the gradient of the energy function~\cite{park2024energy}, which facilitates the generation of multi-concept motions from a single textual description (Figure~\ref{fig:teaser} (d)).

Driven by our exploration, we insightfully find there are two critical limitations of the two spectrums in motion composition: \textbf{\textit{First}}, the latent-aware composition has the problem of text misalignment. \textbf{\textit{Second}}, the semantic-aware composition suffers from foot sliding and motion jitter. To alleviate these issues, we propose a Synergistic Energy Fusion (SEF) strategy by integrating distributions of latent-aware composition, semantic-aware composition, and multi-concept generation.

Taking these innovations together, \textsc{EnergyMoGen} becomes a general and flexible framework for compositional motion generation (our approach can easily generalize to skeleton-based diffusion models, and the results are provided in the Appendix~\ref{sec:supp_add_results}). Empirically, we show impressive results on three datasets across three tasks, \ie, HumanML3D~\cite{guo2022generating} and KIT-ML~\cite{plappert2016kit} for the text-to-motion generation task, MTT~\cite{petrovich24stmc} for compositional motion generation and multi-concept motion generation tasks. We conduct comprehensive experiments for in-depth dissection to reveal the effectiveness of \textsc{EnergyMoGen}. Additionally, we introduce a CompML dataset comprising 5000 compositionally generated motion sequences with textual descriptions. We show that the model's performance can be improved via training on composed motions. 

In summary, our contributions are as follows: (1) Our work is the first attempt to solve the motion composition problem from an energy perspective; (2) We explore two spectrums of energy-based models and propose Synergistic Energy Fusion to address critical issues, \eg, text-misalignment and foot sliding; (3) We introduce two critical operations: negation and the combination of conjunction and negation; (4) We generate a CompML dataset via our motion composition method. We demonstrate that training on CompML can improve performance.

\section{Preliminaries: Energy-Based Models}
\label{sec:background}
In contrast to explicitly modeling the data distribution, Energy-based models (EBMs)~\cite{lecun2006tutorial,zhao2017energybased,liu2024vision} implicitly define the underlying probability density of the distribution through an energy function. For a data sample $\mX$, the probability density can be defined via the Boltzmann distribution:
\begin{equation}
p_\theta(\mX) =  \frac{\exp(-E_\theta(\mX))}{Z(\theta)},
\label{formula:boltzmann}
\end{equation}
where $Z(\theta)=\int\exp(-E_\theta(\mX))d\mX$ is the partition function, and $E_\theta(\mX)$ denotes the energy function, often parameterized as a neural network. However, EBMs struggle to calculate the partition function, making training challenging. Recent endeavors have been devoted to optimizing the Markov Chain Monte Carlo (MCMC) methods via \eg, Langevin dynamics~\cite{du2019implicit,du2023reduce}, contrastive divergence~\cite{du2020improved}, and score-based models~\cite{hyvarinen2005estimation,song2020score}.

In addition, the modern Hopfield networks~\cite{ramsauer2020hopfield,hoover2024energy,hu2024statistical,hu2024provably} interpret the self-attention as EBMs. \cite{ramsauer2020hopfield} introduces an energy function $E(\mX,\xi) = -\alpha^{-1}\log\sum^{N}_{i=1}\exp(\alpha(\mX^{\top}\xi)_i) + \frac{1}{2}\xi^{\top}\xi$ and a update rule $\xi_{new} = \xi - \gamma(\xi -\mX\softmax(\alpha\mX^{\top}\xi))$, where the data sample $\mX$ consists of $N$ stored patterns (key) $\vx_i \in \mathbb{R}^d$, $\xi \in \mathbb{R}^d$ denotes the state vector (query), and $\gamma$ is the update step size. By linearly projecting $\mX$ and $\xi$ into Query ($\mQ$), Key($\mK$), and Value($\mV$), the attention mechanism can be understood as the energy update $\mQ_{new} = \softmax(\alpha\mQ\mK)\mV$, when $\gamma=1$ and $\alpha = 1 / \sqrt{d}$. 
On this basis, Park~\etal~\cite{park2024energy} extends the theory to cross-attention to handle several tasks, \eg, image editing and inpainting.

\begin{figure*}[tp]
\centering
\includegraphics[width=1.0\textwidth]{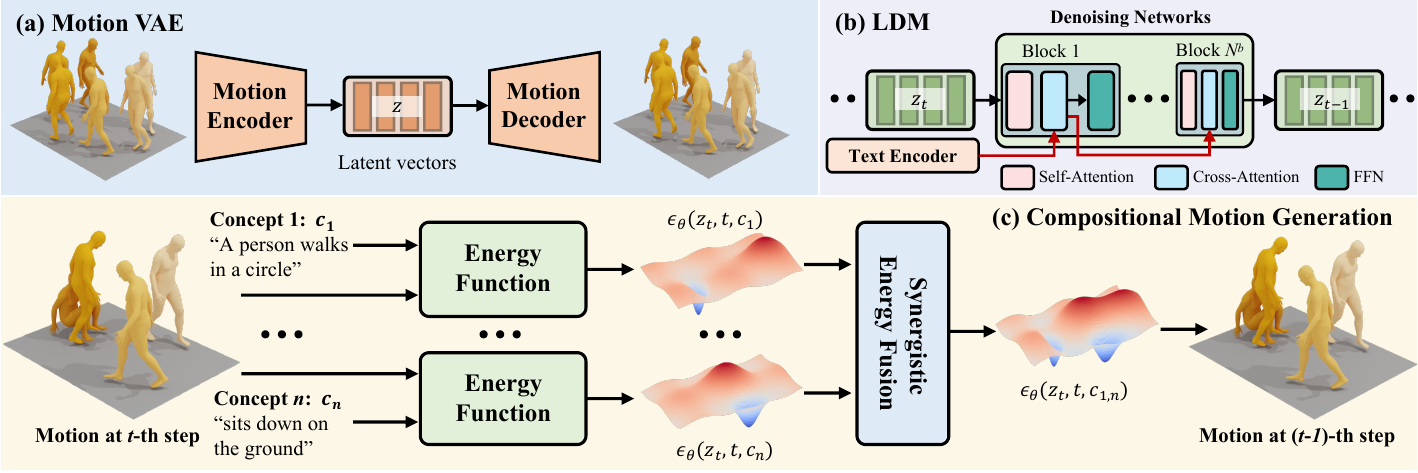}
\vspace{-6mm}
\caption{\textbf{Overview of \textsc{EnergyMoGen}.} (a) Motion Variational Autoencoder (VAE) maps 3D human motion into $N$ latent vectors. (b) We use cross-attention-based transformers as the denoising network in the Latent Diffusion Model (LDM). $N^b$ is the number of transformer layers. Cross-attention can be interpreted as an energy operation that facilitates multi-concept motion generation (Equation~\ref{formula:gradient},~\ref{formula:cross_energy_update}). (c) We explore two spectrums of energy-based models and propose combining them with Synergistic Energy Fusion (SEF). The energy function can be either a denoising network (latent-aware) or a cross-attention (semantic-aware).}
\label{fig:overview}
\vspace{-2mm}
\end{figure*} 

\vspace{-1mm}
\section{Method}
\label{sec:method}
In this section, we introduce our approach, \textsc{EnergyMoGen}. We first present the architecture and formulations of the diffusion model in $\S$~\ref{sec:energymogen}. Next, we discuss the connection between diffusion models and energy-based models, as well as how to compose motions from several concepts through Synergistic Energy Fusion (SEF) in $\S$~\ref{sec:energy}. The overall framework is illustrated in Figure~\ref{fig:overview}.

\subsection{\textsc{EnergyMoGen} for Motion Generation}
\label{sec:energymogen}
We build the architecture of \textsc{EnergyMoGen} upon the motion latent diffusion model~\cite{chen2022mld}, which comprises two transformer-based modules: Motion Variational Autoencoder and Motion Latent Diffusion (in Figure~\ref{fig:overview} (a), (b)).

\vspace{0.5mm}
\noindent\textbf{Motion Variational Autoencoder.}
Given a motion sequence $\mX=[\vx^1, \vx^2, \cdots, \vx^L]$ with $x^l \in \mathbb{R}^{d_m}$, where $L$ and $d_m$ are the length and dimension of motion, we aim to reconstruct the motion via a Variational Autoencoder (VAE). Denoting the encoder and decoder as $\mathcal{E}$ and $\mathcal{D}$, the encoded latent $z=\mathcal{E}(\mX)$, can be sampled from the distribution parameters (mean and variance) using reparameterization trick~\cite{kingma2013auto}, where $z \in \mathbb{R}^{N \times d}$, $N$ and $d$ are the number of vectors and latent dimensions, respectively. With the reconstructed motion $\hat{\mX} = \mathcal{D}(z)$, the VAE is trained to minimize the L1 smooth loss~\cite{Zhang_2023_CVPR} for reconstruction and Kullback-Leibler (KL) divergence between the encoder distribution and normal distribution. 

\vspace{0.5mm}
\noindent\textbf{Motion Latent Diffusion with Energy-Based Cross Attention.} Given the encoded motion latent $z$, a diffusion process aims to model a distribution of $q(z_t|z_{t-1}) = \mathcal{N}(z_t;\sqrt{1-\beta_t}z_{t-1},\beta_tI)$ to obtain a noisy latent $z_t$, where $t \in T$ is the timestep and $0 < \beta \leq 1$.

A denoising autoencoder $\epsilon_\theta$ is employed to predict the underlying score of the distribution $\epsilon \sim \mathcal{N}(0,1)$ based on the text embedding $\vc \in \mathbb{R}^{N^t \times d}$. To optimize $\epsilon_\theta$, we minimize the standard optimization goal, \ie, Mean Square Error (MSE), which can be formulated as follows:
\begin{equation}
\mathbb{E}_{z, \vc, \epsilon, t}\Big[ \Vert \epsilon - \epsilon_\theta(z_t, t, \vc)) \Vert_{2}^{2}\Big].
\label{formula:diffusionloss}
\end{equation}

Building upon the architecture in MLD~\cite{chen2022mld}, we propose to employ cross-attention to incorporate text embeddings and motion latent features.
In light of recent endorsers on the energy-based attention mechanism~\cite{ramsauer2020hopfield,hoover2024energy}, we interpret the cross-attention as an energy-based operation to facilitate multi-concept motion generation. Specifically, $\mK$ and $\mV$ in cross attention is computed as $\mK = \vc W_\mK$ and $\mV = \vc W_\mV$, where $W_\mK, W_\mV \in \mathbb{R}^{d \times d}$. To alleviate the impact of misalignment in multi-concept text embeddings, we focus on refining $\vc$ through adaptive gradient descent (AGD) based on MAP estimation with $p(\mK|\mQ) \propto p(\mQ|\mK)p(\mK)$. We introduce two energy functions, $E(\mK)$ and $E(\mQ|\mK)$, done in the prior work~\cite{park2024energy}. Then, the gradient of the log distribution can be written as:
\begin{equation}
\begin{split}
&\nabla_\mK \log p(\mK|\mQ)=\nabla_\mK E(\mK) - \nabla_\mK E(\mQ|\mK) \\
&=[\underbrace{\texttt{SFM}(\alpha\mK\mQ^{\top})\mQ}_{Attention} - \underbrace{\mathcal{M}(\texttt{SFM}(\mK'))\mK}_{Regularization}]W_\mK,
\end{split}
\label{formula:gradient}
\end{equation}
where \texttt{SFM} and $\mathcal{M}$ are softmax and diagonalization operator, respectively, $\mK'=\frac{1}{2} \sum^{N^t}_{i=1}k_ik_i^{\top}$, $k_i$ denotes the \textit{i}-th row of $\mK$. Finally, $\vc$ is refined with the step size $\gamma$:
\begin{equation}
\hat{\vc} = \vc + \gamma \nabla_\mK \log p(\mK|\mQ).
\label{formula:cross_energy_update}
\end{equation}

By doing so, we can induce the network to focus on the low-energy regions that are relevant to semantics within a multi-concept text. More detailed theoretical analysis can be found in~\cite{park2024energy}.

\subsection{Compositional Human Motion Generation with Energy-Based Diffusion Models}
\label{sec:energy}
With the pre-trained latent motion diffusion model available, we can compose motions by integrating energy functions (Figure~\ref{fig:overview} (c)) in the latent space.

\vspace{1mm}
\noindent\textbf{Bridging Diffusion Models and Energy-Based Models.} During inference, the diffusion model aims to predict the distribution of $p(z_{t-1}|z_t)$ (or $q(z_{t-1}|z_t, z_0)$) to denoise $z_t$ to $z_{t-1}$. By leveraging the Gaussian distribution, we define $p(z_{t-1}|z_t):=\mathcal{N}(z_{t-1};\mu_\theta(z_t, t), \tilde{\beta}_tI)$. With $\mu_\theta(\cdot)$ approximated by $z_t - \epsilon_\theta(z_t, \vc, t)$ for small $\beta_t$, the denoising step can be written as:
\begin{equation}
z_{t-1} = z_t - \epsilon_\theta(z_t, \vc, t) + \mathcal{N}(0, \tilde{\beta}_tI).
\label{formula:diffusion_process}
\end{equation}
In Energy-Based Models (EBMs), the Langevin Dynamics-based MCMC sampling is one of the most popular strategies for the generative process. According to~\cite{du2019implicit,liu2022compositional,du2020compositional,welling2011bayesian}, each gradient descent step in the sampling process is defined as
\begin{equation}
z_{t-1} = z_t - \eta \nabla_z E_\theta(z_t, \vc) + \mathcal{N}(0, \tilde{\beta}_tI),
\label{formula:energy_update}
\end{equation}
which is similar to Equation~\ref{formula:diffusion_process}. By representing the energy function $E_\theta(z_t, \vc)$ using the denoising autoencoder $\epsilon_\theta$, diffusion models can be interpreted as EBMs under the gradient descent step $\eta = 1$, enabling compositional motion generation via composing a set of diffusion models.

\vspace{1mm}
\noindent\textbf{Compositional Motion Generation.} Our goal is to synthesize motions from a set of concepts $\sC = \{\vc_i\}_{i=1}^{n}$ encoded by the text encoder. We introduce compositional operators \textbf{conjunction} and \textbf{negation}. The former seeks to generate motion incorporating all concepts, while the latter proposes the exclusion of certain factors. We use an energy function to model individual concepts $\vc_i$ and assemble them for motion composition. To achieve this, we explore two spectrums of EBMs: \ding{182} \textit{\textbf{latent-aware}} and \ding{183} \textit{\textbf{semantic-aware}}.

\ding{182} For the latent-aware energy-based diffusion model, the probability distribution of conjunction is defined as $p(z|\vc_1,\cdots, \vc_n) \propto p(z) \prod^n_{i=1}p(z|\vc_i)/p(z)$.
Interpret $p(z|\vc_i)$ and $p(z)$ as conditional and unconditional distributions, respectively, we can easily connect $p(z|\vc_1,\cdots, \vc_n)$ to classifier-free guidance~\cite{ho2022classifier}, and the compositional sampling process can be factorized as:
\begin{equation}
\epsilon^l_\theta(z_t,t,\sC) = \epsilon_\theta(z_t,t) + \sum^n_{i=1}w_i^l(\epsilon_\theta(z_t, t, \vc_i) - \epsilon_\theta(z_t,t)),
\label{formula:diffusion_conjunction}
\end{equation}
where $w_i^l$ denotes the weight for \textit{i}-th concept. Similarly, we follow~\cite{du2020compositional,liu2022compositional} to define the probability distribution of negation as $p(z|\vc_i, \text{not} \ \vc_j) \sim \frac{p(z)p(\vc_i|z)}{p(\vc_j|z)}$. Naturally, the composed distribution can be formulated as:
\begin{equation}
\epsilon^l_\theta(z_t,t,\sC) = \epsilon_\theta(z_t,t) + w^l(\epsilon_\theta(z_t, t, \vc_i) - \epsilon_\theta(z_t,t,\vc_j)).
\label{formula:diffusion_negation}
\end{equation}
Through Equation~\ref{formula:diffusion_conjunction} and Equation~\ref{formula:diffusion_negation}, we can generate compositional motions by manipulating latent vectors of pre-trained motion latent diffusion models.

\ding{183} As discussed in $\S$~\ref{sec:energymogen}, cross-attention can be interpreted as an energy-based operation. We utilize this idea to introduce a semantic-aware EBM. Specifically, $\sZ' = \{z'_i\}_{i=1}^n$ is a set of features derived from a cross-attention layer in $\epsilon_\theta$, that corresponds to concepts $\sC$. Due to these output features are unnormalized probability distributions, we compose them by performing a weighted average:
\begin{equation}
\hat{z}' = \frac{1}{\sum^n_{i=1} w^s_i}\sum^n_{i=1} (w^s_i z'_i),
\label{formula:cross_energy_compositon}
\end{equation}
$w^s$ is a hyper-parameter to balance various concepts. $w^s > 0$ indicates a conjunction operation, while $w^s < 0$ suggests a negation operation. We consistently apply this combination across all cross-attention layers, obtaining the output distribution of $\epsilon^s_\theta(z_t,t,\sC)$.

\vspace{1mm}
\noindent\textbf{Synergistic Energy Fusion.} We empirically ($\S$~\ref{sec:exp_indepth}) find that directly employing \ding{182} for compositional motion generation leads to semantic misalignment, while \ding{183} suffers from motion distortion (\eg, foot slides and motion jitter). We propose Synergistic Energy Fusion (SEF) to mitigate these limitations. Given the concepts set $\sC$ and a text embedding $\vc_{1,n}$ of a single textual description with all concepts in $\sC$. At the \textit{t}-th step of the reverse process, we fuse distributions from \ding{182}, \ding{183}, and the single text with multiple concepts as follows:
\begin{equation}
\begin{split}
\hat{\epsilon}_\theta (z_t,t,\sC,\vc_{1,n}) &= \lambda_l \epsilon^l_\theta(z_t,t,\sC) + \lambda_s \epsilon^s_\theta(z_t,t,\sC) \\ & + \lambda_m \epsilon_\theta(z_t,t,\vc_{1,n}),
\end{split}
\label{formula:SEF}
\end{equation}
where $\lambda_l$, $\lambda_s$ and $\lambda_m$ are hyper-parameters with $\lambda_l + \lambda_s + \lambda_m =$ 1, $\hat{\epsilon}_\theta (\cdot)$ is the final score for the next reverse process.

\begin{table*}[t]
    \centering
    \setlength{\tabcolsep}{4pt}
    \resizebox{1.0\linewidth}{!}{

    \begin{tabular}{l c c c c c c c}
    \toprule
    \multirow{2}{*}{Methods}  & \multicolumn{3}{c}{R-Precision $\uparrow$} & \multirow{2}{*}{FID $\downarrow$} & \multirow{2}{*}{MM-Dist $\downarrow$} & \multirow{2}{*}{Diversity $\rightarrow$} & \multirow{2}{*}{MModality $\uparrow$}\\

    \cline{2-4}
    ~ & Top-1 & Top-2 & Top-3 \\
    
    \midrule

        \textbf{Real motion}& \et{0.511}{.003} & \et{0.703}{.003} & \et{0.797}{.002} & \et{0.002}{.000} & \et{2.974}{.008} & \et{9.503}{.065} & -  \\
    \midrule

        MDM~\cite{tevet2022MDM} & \et{0.418}{.005} & \et{0.604}{.001} & \et{0.707}{.004} & \et{0.489}{.025} & \et{3.630}{.023} & \et{9.450}{.066} & \et{2.870}{1.11}  \\ 
        
        MotionDiffuse~\cite{zhang2022motiondiffuse} & \et{0.491}{.001} & \et{0.681}{.001} & \et{0.782}{.001} & \et{0.630}{.001} & \et{3.113}{.001} & \et{9.410}{.049} & \et{1.553}{.042}  \\ 

        M2DM~\cite{kong2023priority} & \et{0.497}{.003} & \et{0.682}{.002} & \et{0.763}{.003} & \et{0.352}{.005} & \et{3.134}{.010} & \et{9.926}{.073} & \etb{3.587}{.072} \\

        Fg-T2M~\cite{wang2023fg} & \et{0.492}{.002} & \et{0.683}{.003} & \et{0.783}{.002} & \et{0.243}{.019} & \et{3.109}{.007} & \et{9.278}{.072} & \et{1.614}{.049}  \\

        ReMoDiffusion~\cite{zhang2023remodiffuse} & \et{0.510}{.005} & \et{0.698}{.006} & \et{\underline{0.795}}{.004} & \etb{0.103}{.004} & \et{2.974}{.016} & \et{9.018}{.075} & \et{1.795}{.043}  \\ 

        FineMoGen~\cite{zhang2023finemogen} & \et{0.504}{.002} & \et{0.690}{.002} & \et{0.784}{.004} & \et{0.151}{.008} & \et{2.998}{.008} & \et{9.263}{.094} & \et{\underline{2.696}}{.079}  \\

        MLD$^*$~\cite{chen2022mld} & \et{0.481}{.003} & \et{0.673}{.003} & \et{0.772}{.002} & \et{0.473}{.013} & \et{3.196}{.010} & \et{9.724}{.082} & \et{2.413}{.079} \\

        GUESS$^*$~\cite{Guess} & \et{0.503}{.003} & \et{0.688}{.002} & \et{0.787}{.002} & \et{\underline{0.109}}{.007} & \et{3.006}{.007} & \et{9.826}{.104} & \et{2.430}{.100} \\

        MotionMamba$^*$~\cite{zhang2025motion}  & \et{0.502}{.003} & \et{0.693}{.002} & \et{0.792}{.002} & \et{0.281}{.009} & \et{3.060}{.058} & \et{9.871}{.084} & \et{2.294}{.058} \\

        \midrule
        
        \textsc{EnergyMoGen}$^*$ & \et{\underline{0.523}}{.003} & \et{\underline{0.715}}{.002} & \etb{0.815}{.002} & \et{0.188}{.006} & \etb{2.915}{.007} & \et{\underline{9.488}}{.099} & \et{2.205}{0.041} \\ 
        \textsc{EnergyMoGen} (CompML)$^*$ & \etb{0.526}{.003} & \etb{0.718}{.003} & \etb{0.815}{.002} & \et{0.176}{.006} & \et{\underline{2.931}}{.007} & \etb{9.500}{.091} & \et{2.270}{0.057} \\ 
    \bottomrule
    \end{tabular}
    }
    \vspace{-1.5mm}
    \caption{\textbf{Comparison with the state-of-the-art diffusion models on the HumanML3D~\cite{guo2022generating} test set.} We repeat the evaluation 20 times for each metric and report the average with a 95\% confidence interval. Bold and underlined indicate the best and second-best results. Methods based on the latent diffusion model are marked with $^*$.}
    \label{tab:HumanML}
    \vspace{-1.5mm}
\end{table*}

\vspace{-1.5mm}
\section{Related Works}
\noindent\textbf{Text-driven Motion Generation.} Text-driven motion generation, a rapidly evolving field, aims to bridge the gap between natural language descriptions and human motion. Earlier works heavily rely on building a text-motion joint latent space~\cite{ahuja2019language2pose,ghosh2021synthesis,tevet2022motionclip}.
However, due to their reliance on autoencoders, these methods are limited in producing high-quality and diverse motions. To address these limitations, various studies have explored VAE-based methods~\cite{guo2022generating,petrovich22temos,petrovich21actor,petrovich2023tmr,SINC:ICCV:2023,TEACH:3DV:2022} to sample motions from Gaussian distribution.
Recently, TM2T~\cite{chuan2022tm2t} and T2M-GPT~\cite{Zhang_2023_CVPR} have made significant progress by projecting human motions into discrete representations via Vector Quantized Variational AutoEncoders (VQ-VAE).
On these bases, some studies focus on optimizing VQ-VAE~\cite{guo2024momask,attt2m,humantomato}, combining with large language models~\cite{zhang2023motiongpt,jiang2023motiongpt}, and exploring the mask language modeling strategy~\cite{pinyoanuntapong2024mmm}.

Diffusion models also show promising results in text-to-motion generation.
Existing diffusion-based methods can be categorized into two groups, depending on the motion representation. The first group typically operates in the original skeleton feature space (composed of joints' position, 6D representation, velocity, \etc). MotionDiffuse~\cite{zhang2022motiondiffuse} is the first effort to leverage diffusion models in text-to-motion generation, while MDM~\cite{tevet2022MDM} is a similar work that uses a Transformer~\cite{vaswani2017attention} with fewer training parameters as the denoising network. Subsequent studies~\cite{dabral2022mofusion,goel2024iterative,han2024amd,ren2023realistic,jin2024act,karunratanakul2023guided,wang2023fg,xie2023omnicontrol,xie2024towards,yang2023synthesizing,zhou2023ude,zhai2023language,yuan2023physdiff} employ carefully designed techniques to improve the performance of the diffusion models, such as retrieval-augmented strategy~\cite{zhang2023remodiffuse}, discrete diffusion~\cite{kong2023priority,lou2023diversemotion}, spatial-temporal attention~\cite{zhang2023finemogen,liu2023spatio}.
Another line of methods applies the diffusion and reverse processes in the continuous latent space. MLD~\cite{chen2022mld} combines the latent diffusion model with motion VAE. GUESS~\cite{gao2024guess} proposes a Gradually Enriching Synthesis strategy based on MLD. MotionMamba~\cite{zhang2025motion} presents a latent diffusion model with Mamba~\cite{gu2023mamba} architecture. 

The most relevant work to ours is MLD~\cite{chen2022mld}, but unlike MLD, we aim at addressing compositional motion generation by rethinking this task from an energy view. We show how our approach enables the latent diffusion model to compose complex motions from simple ones.

\begin{table*}[t]
    \centering
    \resizebox{1.0\linewidth}{!}{

    \begin{tabular}{l c c c c c c c}
    \toprule
    \multirow{2}{*}{Methods}  & \multicolumn{3}{c}{R-Precision $\uparrow$} & \multirow{2}{*}{FID $\downarrow$} & \multirow{2}{*}{MM-Dist $\downarrow$} & \multirow{2}{*}{Diversity $\rightarrow$} & \multirow{2}{*}{MModality $\uparrow$}\\

    \cline{2-4}
    ~ & Top-1 & Top-2 & Top-3 \\
    
    \midrule

        \textbf{Real motion} & \et{0.424}{.005} & \et{0.649}{.006} & \et{0.779}{.006} & \et{0.031}{.004} & \et{2.788}{.012} & \et{11.08}{.097} & -  \\ 
    \midrule

        MDM~\cite{tevet2022MDM} & \et{0.404}{.005} & \et{0.606}{.004} & \et{0.731}{.004} & \et{0.513}{.046} & \et{3.096}{.024} & \et{10.732}{.103} & \et{1.806}{.176} \\
        
        MotionDiffuse~\cite{zhang2022motiondiffuse} & \et{0.417}{.004} & \et{0.621}{.004} & \et{0.739}{.004} & \et{1.954}{.062} & \et{2.958}{.005} & \etb{11.10}{.143} & \et{0.730}{.013}  \\ 

        M2DM~\cite{kong2023priority} & \et{0.416}{.004}  & \et{0.628}{.004} & \et{0.743}{.004} & \et{0.515}{.029} & \et{3.015}{.017} & \et{11.42}{.970} & \etb{3.325}{.370} \\

        Fg-T2M~\cite{wang2023fg} & \et{0.418}{.005} & \et{0.626}{.004} & \et{0.745}{.004} & \et{0.571}{.047} & \et{3.114}{.015} & \et{10.93}{.083} & \et{1.019}{.029}  \\

        ReMoDiffusion~\cite{zhang2023remodiffuse} & \et{0.427}{.014} & \et{0.641}{.004} & \et{\underline{0.765}}{.055} & \etb{0.155}{.006} & \et{2.814}{.012} & \et{10.80}{.105} & \et{1.239}{.028}  \\ 

        FineMoGen~\cite{zhang2023finemogen} & \et{\underline{0.432}}{.006} & \et{\underline{0.649}}{.005} & \etb{0.772}{.006} & \et{\underline{0.178}}{.007} & \et{2.896}{.014} & \et{10.85}{.115} & \et{1.877}{.093}  \\ 
    
        MLD$^*$~\cite{chen2022mld} & \et{0.390}{.008} & \et{0.609}{.008} & \et{0.734}{.007} & \et{0.404}{.027} & \et{3.204}{.027} & \et{10.80}{.117} & \et{2.192}{.071}  \\

        GUESS$^*$~\cite{Guess} & \et{0.425}{.003} & \et{0.632}{.007} & \et{0.751}{.005} & \et{0.371}{.020} & \etb{2.421}{.022} & \et{10.93}{.110} & \et{\underline{2.732}}{.084} \\

        MotionMamba$^*$~\cite{zhang2025motion}  & \et{0.419}{.006} & \et{0.645}{.005} & \et{0.765}{.006} & \et{0.307}{.041} & \et{3.021}{.025} & \et{\underline{11.02}}{.098} & \et{1.678}{.064} \\

        \midrule
        
        \textsc{EnergyMoGen}$^*$ & \etb{0.436}{.006} & \etb{0.651}{.006} & \etb{0.772}{.006} & \et{0.495}{.020} & \et{\underline{2.861}}{.020} & \etb{11.06}{.101} & \et{1.256}{0.024} \\ 
    \bottomrule
    \end{tabular}
    }
    \vspace{-2mm}
    \caption{\textbf{Comparison with the state-of-the-art diffusion models on the KIT-ML~\cite{plappert2016kit} test set.}}
    \vspace{-2mm}
    \label{tab:KIT}
\end{table*}

\vspace{1.0mm}
\noindent\textbf{Compositional Generation.}
Compositional generation has emerged as a powerful paradigm for creating content (\eg, images and motions) that combines multiple components. Our work is related to certain works in compositional image generation. Some approaches aim to develop cross-attention merging algorithms~\cite{feng2023trainingfree,park2024energy}, training strategies~\cite{li2022stylet2i,cong2023attribute,huang2023composer}, and classifier guidance~\cite{shi2023exploring,garipov2023compositional}, to facilitate this task. Another group of works proposes to use EBMs to compose pre-trained generative models~\cite{du2021unsupervised,du2019implicit,liu2021learning,liu2022compositional,park2024energy,du2020compositional}. For example, Du \etal~\cite{du2020compositional} proposed to combine independent trained energy models with several logical operators. Nie \etal~\cite{nie2021controllable} and Du \etal~\cite{du2023reduce} introduced Ordinary Different Equation (ODE) and Markov Chain Monte Carlo (MCMC) to optimize the sampling process, respectively.
In the field of motion generation, TEACH~\cite{TEACH:3DV:2022} and SINC~\cite{SINC:ICCV:2023} are pioneering works for spatial/temporal action composition based on VAE. Subsequent solutions~\cite{shafir2024priormdm,karunratanakul2023gmd,xie2023omnicontrol} manipulate diffusion models to achieve more natural motion and controllability. However, they modify the reverse process in the skeleton space, which is infeasible for adapting to the latent space. 

Drawing inspiration from results in the image domain, we present an energy-based approach for compositional motion generation. In contrast to existing methods, we focus on exploring the motion diffusion model in continuous latent space. Through Synergistic Energy Fusion (SEF), we demonstrate the feasibility of generating high-quality, text-consistent motion by implicitly manipulating latent distributions from multiple components.

\begin{figure*}[tp]
\centering
\includegraphics[width=0.98\textwidth]{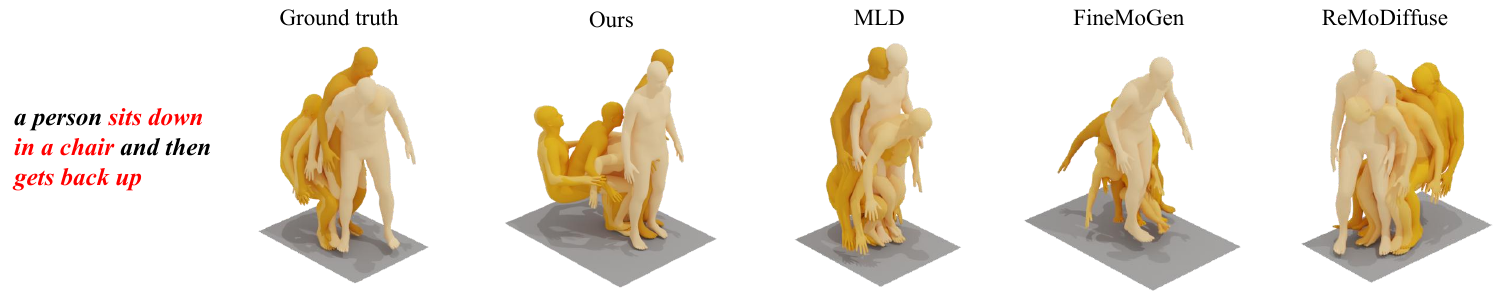}
\vspace{-3mm}
\caption{\textbf{Visual results on the HumanML3D test set.} We compare our approach with MLD~\cite{chen2022mld}, FineMoGen~\cite{zhang2023finemogen}, and ReMoDiffuse~\cite{zhang2023remodiffuse}. Our approach matches the text description better. The motions generated by MLD and FineMoGen are inconsistent with ``\textit{sits down in a chair}'', while ReMoDiffuse fails to generate ``\textit{gets back up}''. More visualization results can be found on the \href{https://jiro-zhang.github.io/EnergyMoGen/}{project page}.}
\vspace{-2mm}
\label{fig:exp_text_to_motion}
\end{figure*} 

\begin{figure*}[tp]
\centering
\includegraphics[width=0.98\textwidth]{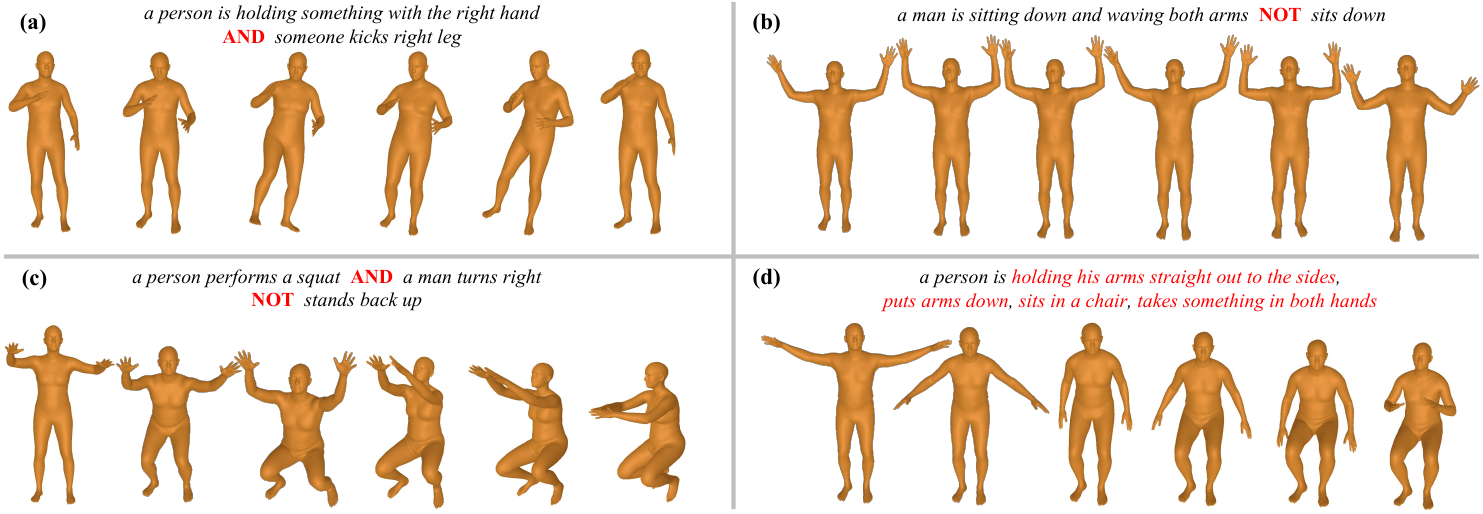}
\vspace{-3mm}
\caption{\textbf{Compositional motion generation.} We use the pre-trained model on HumanML3D~\cite{guo2022generating} for compositional motion generation. Our approach can accurately capture the details in concepts and compose complex motions. (a) conjunction, (b) negation, (c) conjunction + negation, (d) multi-concept generation. More visual results and comparisons can be found on the \href{https://jiro-zhang.github.io/EnergyMoGen/}{project page}.}
\vspace{-2mm}
\label{fig:exp_compositional}
\end{figure*} 

\section{Experiments}
We evaluate our approach on three tasks following prior studies~\cite{guo2022generating,petrovich24stmc}: text-to-motion generation ($\S$~\ref{sec:exp_text2motion}), compositional motion generation ($\S$~\ref{sec:exp_composition}), and multi-concept motion generation ($\S$~\ref{sec:exp_composition}). Furthermore, we offer an in-depth dissection in $\S$~\ref{sec:exp_indepth}. Please kindly note that our approach can be easily generalized to skeleton-based methods. More information on datasets, evaluation metrics, training details, and the results of skeleton-based diffusion models, can be found in the Appendix~\ref{sec:supp_implementation}.
\subsection{Datasets and Evaluation Metrics}
\label{sec:exp_data}
\vspace{1mm}
\noindent\textbf{Datasets.}
Our experiments are conducted on three datasets: HumanML3D~\cite{guo2022generating}, KIT-ML~\cite{plappert2016kit} and MTT~\cite{petrovich24stmc}.
HumanML3D and KIT-ML are used to measure the performance of text-to-motion generation. The composed texts in MTT are used for multi-concept motion generation. We utilize the original three texts and their combination generated by MTT for motion composition.

In addition to these datasets, we build \textbf{CompML}, which consists of 5000 unique motion-text pairs. We randomly select three sentences from a text set (provided in STMC~\cite{petrovich24stmc}) to describe each motion. Then, \textsc{EnergyMoGen} composes motions based on these descriptions. We finetune the pre-trained denoising autoencoder using both CompML and HumanML3D and evaluate it on the HumanML3D test set. We construct this dataset to show that the compositional motion generation can serve as a text-motion data augmentation approach. We also aim to demonstrate that training on motions composed by \textsc{EnergyMoGen} can benefit the text-to-motion generation (see $\S$~\ref{sec:exp_indepth}).

\vspace{1mm}
\noindent\textbf{Evaluation Metrics.}
We use evaluation models from Guo~\etal~\cite{guo2022generating} to measure the performance of text-driven human motion generation. We adopt the same metrics as previous works, including R-Precision, MM-Dist, FID, Diversity, and MModality.
For compositional motion generation and multi-concept motion generation, we follow STMC~\cite{petrovich24stmc} to use R-Precision, TMR-Score, FID, and transition distance for evaluation. 

\begin{table}
\centering
\setlength{\tabcolsep}{3pt}
\resizebox{1.07\linewidth}{!}{
\begin{tabular}{l|cc|cc|cc}
\toprule
\multirow{2}{*}{Methods} & \multicolumn{2}{c|}{R-Presicion} & \multicolumn{2}{c|}{TMR-Score $\uparrow$} & \multirow{2}{*}{\text{FID} $\downarrow$} & Transition \\
& R@1 $\uparrow$ & R@3 $\uparrow$ & M2T & M2M & & distance $\downarrow$ \\

\midrule
\multicolumn{7}{c}{Multi-concept motion generation (single text)} \\
\midrule
MotionDiffuse~\cite{zhang2022motiondiffuse}& 10.9 & 21.3 & 0.558 & 0.546 & 0.621 & 1.9 \\   
MDM~\cite{tevet2022MDM} & 9.5  & 19.7 & 0.556 & 0.549 & 0.666 & 2.5 \\
ReModiffuse~\cite{zhang2023remodiffuse} & 7.4 & 18.3 & 0.531 & 0.534 & 0.699 & 3.3 \\
FineMoGen~\cite{zhang2023finemogen} & 5.4 & 11.7 & 0.504 & 0.533 & 0.948 & 9.4 \\
MLD~\cite{chen2022mld} & 10.5 & 22.3 & 0.559 & 0.552 & 0.685 & 2.4 \\
\midrule
Ours & 12.7  & 25.4 & 0.570 & 0.562 & 0.592 & 2.7 \\
Ours + AGD & 14.0  & 26.3 & 0.570 & 0.560 & 0.587 & 2.7 \\
\midrule
\multicolumn{7}{c}{Compositional motion generation (multiple texts)} \\
\midrule
Ours (latent only) & 9.7  & 19.6 & 0.547 & 0.521 & 0.917 & 1.6 \\
Ours (semantic only) & 15.1 & 27.5 & 0.585 & 0.567 & 0.569 & 2.2 \\
Ours + SEF & 15.9  & 28.0 & 0.591 & 0.567 & 0.604 & 1.6 \\
\bottomrule        
\end{tabular}
}
\vspace{-2mm}
\caption{\textbf{Quantitative comparison on MTT~\cite{petrovich24stmc}}. We compute metrics following STMC~\cite{petrovich24stmc}. `AGD' and `SEF' denote the adaptive gradient descent and synergistic energy fusion. }
\vspace{-3.5mm}
\label{tab:composition}
\end{table}

\subsection{Implementation Details}
During training, the motion encoder, motion decoder, and denoising autoencoder each comprise 9 layers of transformer blocks with a dimension $d=$256. We use 10 additional tokens (mean and various tokens) to sample $N=$5 latent vectors representing the motion. A frozen CLIP ViT-L/14 model is applied to encode the textual descriptions
During inference, we generate motion latent vectors over 50 diffusion steps and then reconstruct them back to motion through the motion decoder. For text-to-motion generation, $\gamma$ is set to 0 for evaluation. Following~\cite{park2024energy}, we split $\gamma$ into $[\gamma_{attn},\gamma_{reg}]$ for compositional and multi-concept motion generation. $[\gamma_{attn},\gamma_{reg}]$ are set to [0.0004, 0.0004] [0.001, 0.002] for two tasks, respectively. 
We evaluate these two tasks on the MTT~\cite{petrovich24stmc} dataset with the model pre-trained on HumanML3D~\cite{guo2022generating}. In addition, $\lambda_{s}$, $\lambda_{l}$, and $\lambda_m$ are set to 0.7, 0.1, and 0.2 for compositional generation evaluation. Ablation studies on hyper-parameters and more training details are provided in the appendix.

\subsection{Experiments on Text-to-Motion Generation}
\label{sec:exp_text2motion}
\noindent\textbf{Quantitative Results.} Table~\ref{tab:HumanML} and Table~\ref{tab:KIT} present the results of \textsc{EnergyMoGen} against famous skeleton-based diffusion~\cite{zhang2023finemogen,zhang2023remodiffuse,wang2023fg} and latent diffusion~\cite{chen2022mld,gao2024guess,zhang2025motion} methods on the HumanML3D and KIT-ML test sets. Our method significantly improves the text-condition consistency and diversity compared with the existing diffusion models while achieving competitive results on fidelity. We demonstrate that leveraging cross-attention with multiple latent vectors is crucial to this enhancement. It's worth noting that our approach, though not initially developed for this specific task, has still demonstrated satisfactory performance.

\vspace{1mm}
\noindent\textbf{Qualitative Results.} Figure~\ref{fig:exp_text_to_motion} shows visual results on HumanML3D~\cite{guo2022generating}. We compare \textsc{EnergyMoGen} with MLD~\cite{chen2022mld} and current state-of-the-art skeleton-based methods: ReMoDiffuse~\cite{zhang2023remodiffuse} and FineMoGen~\cite{zhang2023finemogen}. It can be seen that the motion generated by our approach is more consistent with the textual description. FineMoGen generates some unrealistic frames, and both FineMoGen and MLD fail to generate ``\textit{sits down in a chair}''. The generated motion of ReMoDiffuse~\cite{zhang2023remodiffuse} is not related to ``\textit{gets back up}''.

\begin{figure*}[tp]
\centering
\includegraphics[width=1.0\textwidth]{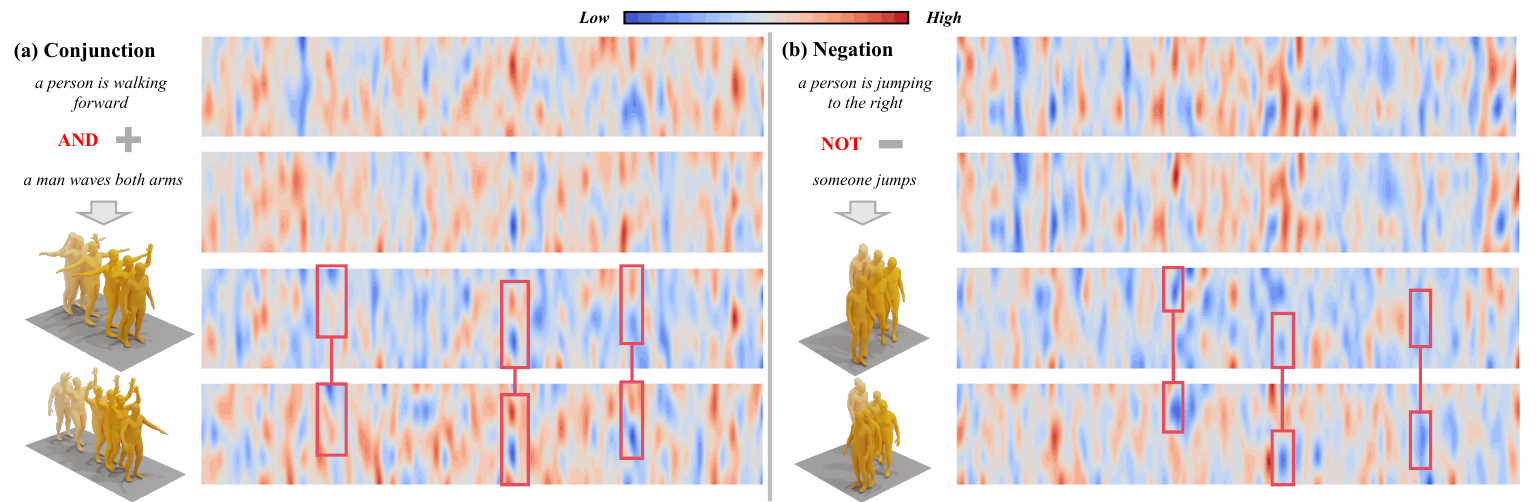}
\vspace{-5mm}
\caption{\textbf{Analysis of the latent distribution.} For a clear illustration, energy distributions are calculated with interpolation and Gaussian smoothing, then visualized as contour maps. Motions in the 4th row are generated from texts, \ie, ``a person is walking forward and waving both arms'' and ``a person is walking to the right'', which is created by composing the multiple concepts into a single text via (a) Conjunction and (b) Negation. Similar regions are highlighted in \textcolor{red}{red}.}
\label{fig:exp_energy_visual}
\end{figure*} 

\subsection{Experiments on Motion Composition}
\label{sec:exp_composition}
\noindent\textbf{Quantitative Results.}
In Table~\ref{tab:composition}, we perform a comprehensive comparison against two divergent groups of diffusion models (\ie, skeleton-based and latent-based) on the MTT dataset. We empirically find that the motion jitter issue in FineMoGen~\cite{zhang2023finemogen} and ReMoDiffuse~\cite{zhang2023remodiffuse} hinders their performance on this task. Our approach outperforms both types of rivals on multi-concept and compositional motion generation tasks. Precisely, \textsc{EnergyMoGen} with adaptive gradient descent of text embeddings (in Equation~\ref{formula:cross_energy_update}) outperforms previous state-of-the-art methods on R-Precision, TMR-Score, and FID. With synergistic energy fusion, \textsc{EnergyMoGen} significantly exhibits significant performance advantages over all prior methods. 

\vspace{1mm}
\noindent\textbf{Qualitative Results.} In Figure~\ref{fig:exp_compositional}, we show visual results of compositional motion generation in four settings: (a) conjunction, (b) negation, (c) conjunction + negation, and (d) multi-concept generation. Our approach can generate high-quality motions that are consistent with multiple concepts.

\subsection{In-Depth Dissection of \textsc{EnergyMoGen}}
\label{sec:exp_indepth}
\noindent\textbf{Key Component Analysis.} We first investigate the key components in our \textsc{EnergyMoGen}.

\textit{\textbf{Q1: Does the adaptive gradient descent alleviate the issue of text inconsistency in multi-concept motion generation?} Yes.} As shown in Table~\ref{tab:composition}, our approach with adaptive updating of the textual embedding aligns better with textual descriptions. It improves the R-Precision by 1.3\% and 0.9\% for Top1 and Top3 accuracy, respectively. Meanwhile, we achieve better FID and comparable results on the TMR-Score and transition distance. The ablation study of $[\gamma_{attn},\gamma_{reg}]$ is provided in the appendix.

\textit{\textbf{Q2: What roles do latent-aware and semantic-aware EBMs play in motion composition?}} To answer this question, we conduct ablative experiments on MTT~\cite{petrovich24stmc}, results are shown in Table~\ref{tab:composition}. One can figure out that the semantic-aware EBM significantly enhances the text-motion consistency, but it suffers from foot slides and motion jitter. Although the latent-aware method performs poorly on R-Precision and TMR Score, it clearly decreases the transition distance. We notice that Synergistic Energy Fusion effectively alleviates these issues by leveraging the complementary strengths of the two spectrums.
We present ablation studies on $\lambda_{s}$, $\lambda_{l}$, and $\lambda_m$ in the appendix.

\textit{\textbf{Q3: How does \textsc{EnergyMoGen} compose complex motions through energy?}} We study the underlying mechanism of our approach from the perspectives of \textbf{conjunction} and \textbf{negation}. We visualize the energy distributions of motion latent representations generated from the denoising autoencoder in Figure~\ref{fig:exp_energy_visual}. In concept conjunction, we add the two distributions together, and the contour maps from compositional generation (the 3rd row) and multi-concept generation (the last row) demonstrate consistent high- and low-energy regions. In concept negation, we input a text of ``\textit{a person is walking to the right}'' based on the motion generated by composing ``\textit{a person is jumping to the right}'' and ``\textit{someone jumps}''. Despite the absence of ``walking'' in the original concepts, the energy distributions of the two motions (the 3rd and 4th rows) still exhibit similar regions.
Similar energy regions are highlighted in \textcolor{red}{red}.
Such results potentially explain the effectiveness of \textsc{EnergyMoGen}.

\vspace{1mm}
\noindent\textbf{Experiments on CompML.} We finetune the pretrained denoising network using both HumanML3D and CompML. The results are illustrated in Table~\ref{tab:HumanML}. We observe that the model's performance improves with additional training data from CompML. Our experiments demonstrate that compositional motion generation can be effectively employed as a data augmentation technique. Furthermore, such results also indicate that the motions generated by our approach are of sufficient quality to be used for data augmentation.
\section{Conclusion}
In this paper, we present a framework for compositional human motion generation, aiming at composing complex motions from a set of simple concepts. We explore an energy-based regime where the denoising autoencoder and cross-attention are interpreted as energy functions for latent-aware and semantic-aware composition, and energy distributions are combined in latent space through Synergistic Energy Fusion. Our approach consistently shows superior performance across various tasks on several commonly used benchmarks.
Extensive investigations demonstrate that our approach is general, effective and interpretable. Moreover, we build a dataset and suggest motion composition as a data augmentation technique, which can bring additional improvement to text-to-motion generation. 
\section*{Acknowledgments}
This work is supported by the National Natural Science Foundation of China (62472381), Fundamental Research Funds for the Zhejiang Provincial Universities (226-2024-00208), and the Earth System Big Data Platform of the School of Earth Sciences, Zhejiang University.

{
    \small
    \bibliographystyle{ieeenat_fullname}
    \bibliography{main}
}

\clearpage
\appendix
\vspace*{1em}{\centering\large\bf%
Appendix
\vspace*{1.5em}}

In this appendix, we present:
\begin{itemize}
    
    \item Section~\ref{sec:supp_implementation}: Training details of \textsc{EnergyMoGen}.\\

    \item Section~\ref{sec:supp_add_results}: Additional results of skeleton-based diffusion models.\\

    \item Section~\ref{sec:supp_ablation_composition}: Ablations on $\lambda_{s}$, $\lambda_{l}$, and $\lambda_m$ for Synergistic Energy Fusion.\\

    \item Section~\ref{sec:supp_nb_latent}: Ablations on the number of latent vectors in motion VAE.\\
    
    \item Section~\ref{sec:supp_hyper}: Ablation study of hyper-parameters in energy-based cross-attention.\\

    \item Section~\ref{sec:supp_inference_time}: Results of inference time. \\

    \item Section~\ref{sec:supp_foot_sliding}: Evaluation of foot sliding. \\
    
    \item Section~\ref{sec:supp_energy_visual}: More visual results of energy distributions.\\

    \item Section~\ref{sec:supp_details_data_eval}: More details on datasets and evaluation metrics.\\

    \item Section~\ref{sec:limitations}: Limitation and failure case
    
\end{itemize}
\begin{table*}[t]
    \centering
    \setlength{\tabcolsep}{4pt}
    \resizebox{1.0\linewidth}{!}{

    \begin{tabular}{l c c c c c c c}
    \toprule
    \multirow{2}{*}{Methods}  & \multicolumn{3}{c}{R-Precision $\uparrow$} & \multirow{2}{*}{FID $\downarrow$} & \multirow{2}{*}{MM-Dist $\downarrow$} & \multirow{2}{*}{Diversity $\rightarrow$} & \multirow{2}{*}{MModality $\uparrow$}\\

    \cline{2-4}
    ~ & Top-1 & Top-2 & Top-3 \\
    
    \midrule

        \textbf{Real motion}& \et{0.511}{.003} & \et{0.703}{.003} & \et{0.797}{.002} & \et{0.002}{.000} & \et{2.974}{.008} & \et{9.503}{.065} & -  \\
    \midrule

        MDM~\cite{tevet2022MDM} & \et{0.418}{.005} & \et{0.604}{.001} & \et{0.707}{.004} & \et{0.489}{.025} & \et{3.630}{.023} & \etb{9.450}{.066} & \etb{2.870}{1.11}  \\ 
        
        MotionDiffuse~\cite{zhang2022motiondiffuse} & \et{0.491}{.001} & \et{0.681}{.001} & \et{0.782}{.001} & \et{0.630}{.001} & \et{3.113}{.001} & \et{\underline{9.410}}{.049} & \et{1.553}{.042}  \\ 

        ReMoDiffusion~\cite{zhang2023remodiffuse} & \et{\underline{0.510}}{.005} & \et{\underline{0.698}}{.006} & \et{\underline{0.795}}{.004} & \etb{0.103}{.004} & \et{\underline{2.974}}{.016} & \et{9.018}{.075} & \et{1.795}{.043}  \\ 

        FineMoGen~\cite{zhang2023finemogen} & \et{0.504}{.002} & \et{0.690}{.002} & \et{0.784}{.004} & \et{0.151}{.008} & \et{2.998}{.008} & \et{9.263}{.094} & \et{\underline{2.696}}{.079}  \\ 
        \midrule
        \textsc{EnergyMoGen} (skeleton) & \etb{0.528}{.003} & \etb{0.718}{.003} & \etb{0.810}{.002} & \et{\underline{0.139}}{.007} & \etb{2.902}{.010} & \et{9.386}{.078} & \et{2.549}{0.104} \\ 
    \bottomrule
    \end{tabular}
    }
    \caption{\textbf{Comparison with the state-of-the-art diffusion models on the HumanML3D~\cite{guo2022generating} test set.} We repeat the evaluation 20 times for each metric and report the average with a 95\% confidence interval. Bold and underlined indicate the best and second-best results.}
    \label{tab:supp_HumanML}
\end{table*}

\begin{table*}[t]
    \centering
    \setlength{\tabcolsep}{4pt}
    \resizebox{0.75\linewidth}{!}{

    \begin{tabular}{l c c c c}
    \toprule
    Methods & R-Precision $\uparrow$ & FID $\downarrow$ & Diversity $\rightarrow$ & MM-Dist $\downarrow$ \\
    \midrule
    Ground Truth & 0.80 & 1.6 $\times$ 10$-3$ & 9.62 & 2.96 \\ \midrule
    PriorMDM~\cite{shafir2024priormdm} (Double take) & 0.59 & 0.60 & \underline{9.50} & 5.61 \\
    PriorMDM~\cite{shafir2024priormdm} (First take) & 0.59 & 1.00 & 9.46 & 5.63 \\
    MotionDiffuse~\cite{zhang2025motion} & 0.62 & 1.76 & 8.55 & 5.40 \\
    ReMoDiffuse~\cite{zhang2023remodiffuse} & \underline{0.64} & \textbf{0.40} & 9.35 & \underline{5.24}  \\
    FineMoGen~\cite{shafir2024priormdm} & \underline{0.64} & 0.45 & 9.23 & 5.27 \\ \midrule
    Ours & \textbf{0.67} & \underline{0.43} & \textbf{9.52} & \textbf{5.22} \\ 

    \bottomrule
    \end{tabular}
    }
    \caption{\textbf{Quantitative results on the HumanML3D~\cite{guo2022generating} test set.} R-Presicion denotes Top-3 accuracy. Bold and underlined indicate the best and second-best results.}
    \label{tab:supp_temporal}
\end{table*}

\section{Implementation Details}
\label{sec:supp_implementation}
We first provide training details of \textsc{EnergyMoGen}. For Motion VAE, both the encoder $\mathcal{E}$ and decoder $\mathcal{D}$ comprise 9 layers of transformer blocks with a dimension $d=$256. We use 10 additional tokens (mean and various tokens) to sample $N=$5 latent vectors representing the motion. We use an AdamW optimizer with a batch size of 1024. We train 300K iterations in total, and the learning rate changes from 0.0001 to 0.00001 after 200K iterations. The weights of reconstruction loss and KL loss are set to 1 and 0.0001.
As for the latent diffusion, we apply a frozen CLIP ViT-L/14 to encode the textual descriptions. Regarding the denoising autoencoder, we use a 9-layer transformer with a dimension of 256. To acquire an accurate mapping from textual data to latent vectors during training, $\gamma$ is initialized with 0. We utilize the AdamW optimizer to train the model with a batch size of 512, with an initial learning rate of 0.0001 for 200K iterations and decayed to 0.00001 for another 100K iterations. The diffusion model is learned using classifier-free guidance~\cite{ho2022classifier} with an unconditional score estimation rate of 10\%. 
For experiments on CompML, we only finetune the latent diffusion model for 100K iterations in total with a learning rate of 0.00005.

For the skeleton-based approach, we use an 8-layer transformer with a dimension of 512. We follow~\cite{zhang2022motiondiffuse} to train the model using the Adam optimizer with a batch size of 1024. We train 8000 epochs in total and employ the CosineAnnealing learning policy with the learning rate from 0.0002 to 0.00002. 

\section{Additional Results of Skeleton-Based Diffusion Models}
\label{sec:supp_add_results}
\subsection{Text-to-Motion Generation}
We conduct experiments on HumanML3D~\cite{guo2022generating} to evaluate the performance of text-to-motion generation. We use evaluation models from Guo~\etal~\cite{guo2022generating} and use the same metrics.
The training details of skeleton-based \textsc{EnergyMoGen} are provided in Section~\ref{sec:supp_implementation}. Experimental results are shown in Table~\ref{tab:supp_HumanML}. Our approach outperforms current state-of-the-art skeleton-based methods, \ie, ReMoDiffuse~\cite{zhang2023remodiffuse} and FineMoGen~\cite{zhang2023finemogen} on R-Precision, Diversity, and MM-Dist, while achieving comparable results on FID and MModality.

\subsection{Motion Temporal Composition}
Following FineMoGen~\cite{zhang2023finemogen} and PriorMDM~\cite{shafir2024priormdm}, we use the motion temporal composition task to measure the compositional capacity of our approach. We perform latent-aware composition to tackle this task. 

Specifically, denoting $\vc_1$ and $\vc_2$ as two concepts. $M_{\vc_1}^t \in \mathbb{R}^{N_1 \times d_m}$, $M_{\vc_2}^t \in \mathbb{R}^{N_2 \times d_m}$ denote predicted scores corresponding to two concepts at $t$-th step, $N_i$ is the motion length, and $d_m$ is the dimension of motions. $M_3^t \in \mathbb{R}^{N' \times d_m}$ indicates the overlapping part, where $N'$ is the number of interval frame. Each reverse process can be formulated as:
\begin{equation}
\begin{split}
M^t_{\vc_1, \vc_2} =& M_{\vc_1}^t[:N_1-N'] \oplus (M_{\vc_1}^t[N_1-N':] + \\
&M_{\vc_2}^t[:N'] - M_3^t) \oplus M_{\vc_2}^t[N':], \\
\end{split}
\label{formula:temporal}
\end{equation}
where $M^t_{\vc_1, \vc_2}$ is the final score at $t$-th step, $\oplus$ is the concatenate operation. We conduct experiments on the HumanML3D dataset, and the results are shown in Table~\ref{tab:supp_temporal}. We implement MotionDiffuse~\cite{zhang2022motiondiffuse}, ReMoDiffuse~\cite{zhang2023remodiffuse}, and FineMoGen~\cite{zhang2023finemogen} using the ``first take'' from PriorMDM~\cite{shafir2024priormdm}. Our approach is implemented based on Equation~\ref{formula:temporal}, and exhibits performance advantages compared with previous methods. We provide visual comparisons with PriorMDM, which can be found on the \href{https://jiro-zhang.github.io/EnergyMoGen/}{project page}.

\subsection{Multi-Concept Motion Generation}
In Table~\ref{tab:supp_multi}, we show quantitative results on the MTT~\cite{petrovich24stmc} dataset. Our approach without Adaptive Gradient Descent (AGD) yields results that are competitive with existing state-of-the-art methods. By combining AGD, our approach achieves superior performance on R-Precision, TMR-Score, and Transition distance.

\begin{table*}
\centering
\setlength{\tabcolsep}{6pt}
\resizebox{0.82\linewidth}{!}{
\begin{tabular}{l|cc|cc|cc}
\toprule
\multirow{2}{*}{Methods} & \multicolumn{2}{c|}{R-Presicion} & \multicolumn{2}{c|}{TMR-Score $\uparrow$} & \multirow{2}{*}{\text{FID} $\downarrow$} & Transition \\
& R@1 $\uparrow$ & R@3 $\uparrow$ & M2T & M2M & & distance $\downarrow$ \\
\midrule
MotionDiffuse~\cite{zhang2022motiondiffuse}& \underline{10.9} & 21.3 & \underline{0.558} & 0.546 & \textbf{0.621} & \textbf{1.9} \\   
MDM~\cite{tevet2022MDM} & 9.5  & 19.7 & 0.556 & \underline{0.549} & 0.666 & 2.5 \\
ReModiffuse~\cite{zhang2023remodiffuse} & 7.4 & 18.3 & 0.531 & 0.534 & 0.699 & 3.3 \\
FineMoGen~\cite{zhang2023finemogen} & 5.4 & 11.7 & 0.504 & 0.533 & 0.948 & 9.4 \\
\midrule
\textsc{EnergyMoGen} (skeleton) & \textbf{11.5}  & \underline{22.6} & 0.550 & \underline{0.549} & 0.670 & \underline{2.2} \\
\textsc{EnergyMoGen} (skeleton) + AGD & \textbf{11.5}  & \textbf{24.4} & \textbf{0.560} & \textbf{0.552} & \underline{0.643} & \textbf{1.9} \\
\bottomrule        
\end{tabular}
}
\caption{\textbf{Quantitative comparison of skeleton-based diffusion on MTT~\cite{petrovich24stmc}}. We compute metrics following STMC~\cite{petrovich24stmc}. `AGD' denotes the adaptive gradient descent.}
\label{tab:supp_multi}
\end{table*}

\begin{table*} %
\centering
\setlength{\tabcolsep}{8pt}
\resizebox{0.75\linewidth}{!}{
\begin{tabular}{ccc|cccc|cc}
\toprule
& & & \multicolumn{4}{c|}{Per-crop semantic correctness} & \multicolumn{2}{c}{Realism} \\
$\lambda_{l}$ & $\lambda_{s}$ & $\lambda_m$ & \multirow{2}{*}{R@1 $\uparrow$} & \multirow{2}{*}{R@3 $\uparrow$} & \multicolumn{2}{c|}{TMR-Score $\uparrow$} & \multirow{2}{*}{\text{FID} $\downarrow$} & Transition \\
& & & & & M2T & M2M & & distance $\downarrow$ \\

\midrule
1.0 & 0.0 & 0.0 & 9.7  & 19.6 & 0.547 & 0.521 & 0.917 & 1.6 \\
0.0 & 1.0 & 0.0 & \underline{15.1} & \underline{27.5} & 0.585 & \textbf{0.567} & \textbf{0.569} & 2.2 \\
0.0 & 0.0 & 1.0 & 12.7  & 25.4 & 0.570 & 0.562 & \underline{0.592} & 2.7 \\
\midrule
0.5 & 0.5 & 0.0 & 13.3 & 25.5 & 0.584 & 0.551 & 0.740 & \textbf{1.4} \\
0.4 & 0.6 & 0.0 & 12.7 & 25.0 & 0.584 & 0.555 & 0.730 & \textbf{1.4} \\
0.3 & 0.7 & 0.0 & 13.4  & 26.4 & 0.589 & 0.559 & 0.694 & \textbf{1.4} \\
0.2 & 0.8 & 0.0 & 13.6 & 26.9 & \underline{0.590} & 0.558 & 0.668 & \underline{1.5} \\
0.1 & 0.9 & 0.0 & 14.2  & \underline{27.5} & 0.587 & \underline{0.563} & 0.613 & 1.7 \\
\midrule
0.3 & 0.4 & 0.3 & 14.5  & 27.1 & 0.588 & 0.560 & 0.669 & 1.6 \\
0.2 & 0.5 & 0.3 & 14.4  & 26.9 & \underline{0.590} & \underline{0.563} & 0.628 & 1.6 \\
0.1 & 0.7 & 0.2 & \textbf{15.7} & \textbf{28.0} & \textbf{0.591} & \textbf{0.567} & 0.604 & 1.6 \\
0.1 & 0.8 & 0.1 & 14.9  & 26.7 & 0.587 & \underline{0.563} & 0.615 & 1.6 \\

\bottomrule        
\end{tabular}
}
\caption{\textbf{Ablation of hyper-parameters in Synergistic Energy Fusion on MTT~\cite{petrovich24stmc}.} We find that as the weight of $\lambda_s$ increases, the results align more closely with the text (R-Precision and TMR-Score), while larger weights for $\lambda_l$ produce smoother motions (Transition distance).
}
\label{tab:supp_ablation_sef}
\end{table*}

\begin{figure}[t]
 \centering
 \begin{subfigure}[b]{0.497\textwidth}
    \includegraphics[width=\linewidth]{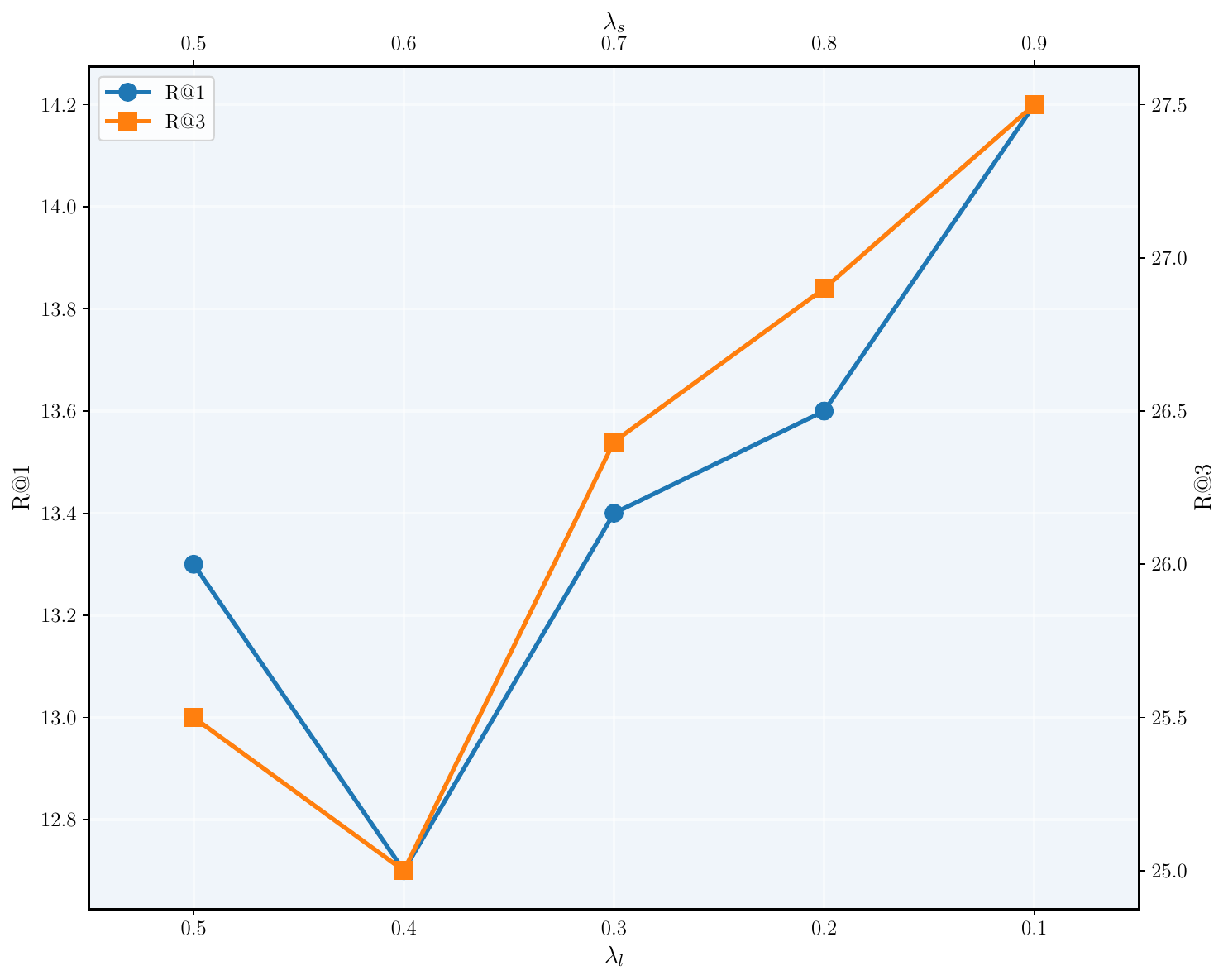}
    \caption{}
    \label{fig:diversity-teaser}
  \end{subfigure}
  \hfill 
  \begin{subfigure}[b]{0.49\textwidth}
    \includegraphics[width=\linewidth]{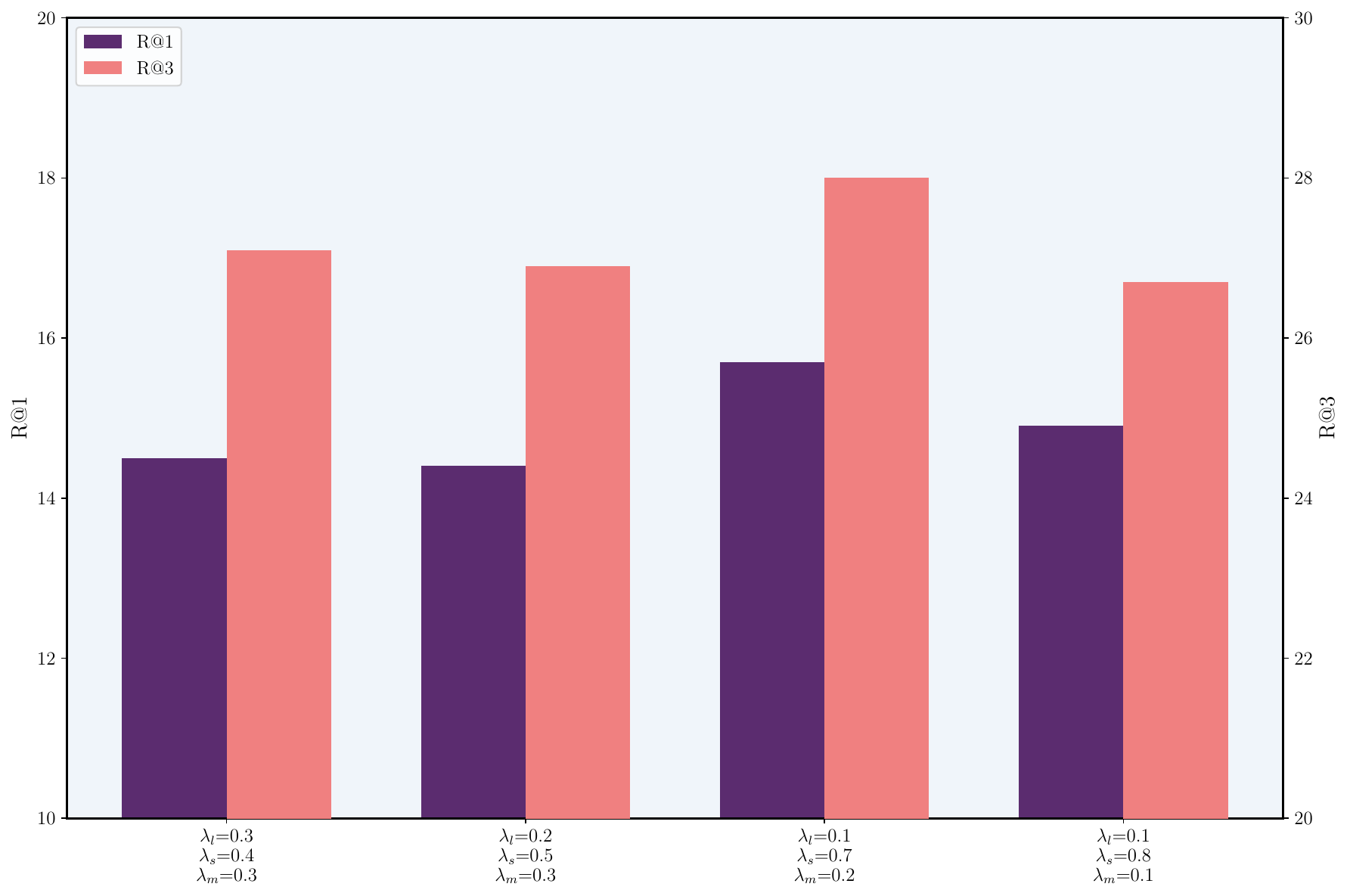}
    \caption{}
    \label{fig:fid-teaser}
  \end{subfigure}
  \caption{\textbf{Ablation of hyper-parameters in Synergistic Energy Fusion on MTT~\cite{petrovich24stmc}.}}
  \label{fig:intuitive}
\end{figure}

\section{Ablations on Synergistic Energy Fusion}
\label{sec:supp_ablation_composition}
We show the effect of hyper-parameters $\lambda_{l}$, $\lambda_{s}$, and $\lambda_m$ in Table~\ref{tab:supp_ablation_sef}.
The results in the first three rows correspond to ``Ours (latent only)'', ``Ours (semantic only)'', and ``Ours'' in Table~\textcolor{red}{3} of the main paper, respectively.

Then we conduct ablative experiments on the weights of the two spectra of the energy-based model (latent-aware and semantic-aware), as shown in the middle five rows (Table~\ref{tab:supp_ablation_sef}). We find that as the weight of $\lambda_s$ increases, the results align more closely with the text (R-Precision and TMR-Score), while larger weights for $\lambda_l$ produce smoother motions (Transition distance).

Meanwhile, we also demonstrate that combining multi-concept motion generation can further improve the performance, as shown in the last 4 rows. It can be seen that Synergistic Energy Fusion with $\lambda_{l} = 0.1$, $\lambda_{s}=0.7$, and $\lambda_m=0.2$ achieves best performance. 

Note that a more intuitive comparison can be found in Figure~\ref{fig:intuitive}.

\section{Ablations on the Number of Latent Vectors \textit{N} in Motion VAE}
\label{sec:supp_nb_latent}
The results are provided in Table~\ref{tab:supp_nb_latent}. For reconstruction, 7 latent vectors achieve the best results. However, it increases the difficulty of latent diffusion models in training. Using 5 latent vectors to represent the motion obtains the best text-to-motion generation performance.

\begin{table*}[t]
    \centering
    \setlength{\tabcolsep}{8pt}
    \resizebox{0.85\linewidth}{!}{

    \begin{tabular}{c c c c c c c}
    \toprule
    \multirow{2}{*}{$N$}  & \multicolumn{3}{c}{R-Precision $\uparrow$} & \multirow{2}{*}{FID $\downarrow$} & \multirow{2}{*}{MM-Dist $\downarrow$} & \multirow{2}{*}{Diversity $\rightarrow$}\\

    \cline{2-4}
    ~ & Top-1 & Top-2 & Top-3 \\
    \midrule
    \multicolumn{7}{c}{Reconstruction} \\
    \midrule
        1 & \et{0.493}{.002} & \et{0.681}{.002} & \et{0.787}{.003} & \et{0.170}{.001} & \et{3.160}{.015} & \etb{9.589}{.074} \\ 
        3 & \et{0.501}{.002} & \et{0.696}{.002} & \et{0.792}{.004} & \et{0.117}{.000} & \et{3.037}{.007} & \et{9.621}{.091} \\ 
        5 & \et{0.508}{.003} & \et{0.700}{.003} & \et{0.795}{.002} & \et{0.080}{.000} & \et{3.004}{.008} & \et{9.620}{.098} \\ 
        7 & \etb{0.513}{.002} & \etb{0.704}{.003} & \etb{0.797}{.002} & \etb{0.022}{.000} & \etb{2.984}{.009} & \et{9.603}{.085} \\ 
    \midrule
    \multicolumn{7}{c}{Generation} \\
    \midrule
        1 & \et{0.498}{.003} & \et{0.686}{.004} & \et{0.791}{.004} & \et{0.424}{.009} & \et{3.085}{.009} & \et{9.705}{.097} \\ 
        3 & \et{0.523}{.004} & \et{0.712}{.002} & \et{0.814}{.002} & \et{0.418}{.025} & \et{2.946}{.009} & \et{9.443}{.136} \\ 
        5 & \etb{0.523}{.003} & \etb{0.715}{.002} & \etb{0.815}{.002} & \etb{0.188}{.006} & \etb{2.915}{.007} & \etb{9.488}{.099} \\ 
        7 & \et{0.514}{.004} & \et{0.713}{.005} & \et{0.813}{.003} & \et{0.291}{.006} & \et{2.938}{.012} & \et{9.456}{.130} \\ 
    
    \bottomrule
    \end{tabular}
    }
    \caption{\textbf{Study on the number of latent vectors in motion VAE on the HumanML3D~\cite{guo2022generating} test set.}}
    \label{tab:supp_nb_latent}
\end{table*}

\section{Ablation Study of Hyper-parameters in Cross-Attention}
\label{sec:supp_hyper}
We investigate the impact of $\gamma_{attn}$ and $\gamma_{reg}$ We follow~\cite{park2024energy} to split $\gamma$ into attention step size $\gamma_{attn}$ and regularization step size $\gamma_{reg}$ for compositional and multi-concept motion generation in energy-based cross-attention, and the results are presented in Table~\ref{tab:supp_ablation_cross}. We notice that $\gamma_{attn},\gamma_{reg} >= 0.1$ significantly degrades the performance, and $\gamma_{attn},\gamma_{reg} = [0.001, 0.002]$ achieves best results on MTT.

\begin{table*} %
\centering
\setlength{\tabcolsep}{8pt}
\resizebox{0.7\linewidth}{!}{
\begin{tabular}{cc|cccc|cc}
\toprule
& & \multicolumn{4}{c|}{Per-crop semantic correctness} & \multicolumn{2}{c}{Realism} \\
$\gamma_{attn}$ & $\gamma_{reg}$ & \multirow{2}{*}{R@1 $\uparrow$} & \multirow{2}{*}{R@3 $\uparrow$} & \multicolumn{2}{c|}{TMR-Score $\uparrow$} & \multirow{2}{*}{\text{FID} $\downarrow$} & Transition \\
& & & & M2T & M2M & & distance $\downarrow$ \\

\midrule
0.0 & 0.0 & 12.7  & 25.4 & \textbf{0.570} & \textbf{0.562} & 0.592 & 2.7 \\
0.1 & 0.2 & 1.4  & 4.2  & 0.498 & 0.495 & 1.083 & 6.1 \\
0.01 & 0.02 & 9.1  & 20.1 & 0.551 & 0.547 & 0.623 & 2.9 \\
0.005 & 0.01 & 6.7 & 15.3 & 0.531 & 0.523 & 0.806 & \textbf{1.9} \\
0.005 & 0.005 & 13.8  & 25.6 & \textbf{0.570} & 0.558 & 0.591 & 2.7 \\
0.001 & 0.002 & \textbf{14.0}  & \textbf{26.3} & \textbf{0.570} & 0.560 & \textbf{0.587} & 2.7 \\
\bottomrule        
\end{tabular}
}
\caption{\textbf{Ablation of step size in Adaptive Gradient Descent on MTT~\cite{petrovich24stmc}.}
}
\label{tab:supp_ablation_cross}
\end{table*}

\section{Inference Time}
\label{sec:supp_inference_time}
Since our method, like most others, is based on Transformer, we compare its inference time with SOTA Transformer-based diffusion models in Table~\ref{tab:average_time}.
\begin{table}[h]
    \centering\setlength{\tabcolsep}{4pt}
    \resizebox{1.0\columnwidth}{!}{
    \begin{tabular}{l|ccc|c}
    \toprule
        Methods & FineMoGen & MLD & GUESS & \textsc{EnergyMoGen}  \\ \midrule
        AIT (s) & 2.54 & \textbf{0.21} & 1.79 & 0.66 \\ \bottomrule
    \end{tabular}
    }
    \vspace{-2mm}
    \caption{\textbf{Inference time}. AIT (s) denotes the Average inference time per sentence in seconds.}
    \vspace{-2mm}
    \label{tab:average_time}
\end{table}

\section{Evaluation of Foot Sliding}
\label{sec:supp_foot_sliding}
Physical Foot Contact score (PFC), proposed in EDGE~\cite{tseng2022edge}, is used to evaluate the foot sliding problem. We provide a PFC comparison on MTT in Table~\ref{tab:evluation_slide}, demonstrating the effectiveness of the proposed Synergistic Energy Fusion.

\begin{table}[t]
    \centering\setlength{\tabcolsep}{4pt}
    \resizebox{\columnwidth}{!}{
    \begin{tabular}{l|ccc|c}
    \toprule
        Methods & multi-concept & latent-only & semantic-only & Ours SEF  \\ \midrule
        PFC $\downarrow$ & 0.61 & 0.54 & 1.05 & \textbf{0.51} \\ \bottomrule
    \end{tabular}
    }
    \vspace{-2mm}
    \caption{\textbf{Evaluation of Foot Sliding}. `PFC' denotes the Physical Foot Contact score.}
    \vspace{-2mm}
    \label{tab:evluation_slide}
\end{table}

\section{Energy Distribution Visualization}
\label{sec:supp_energy_visual}
\begin{figure*}[tp]
\centering
\includegraphics[width=\textwidth]{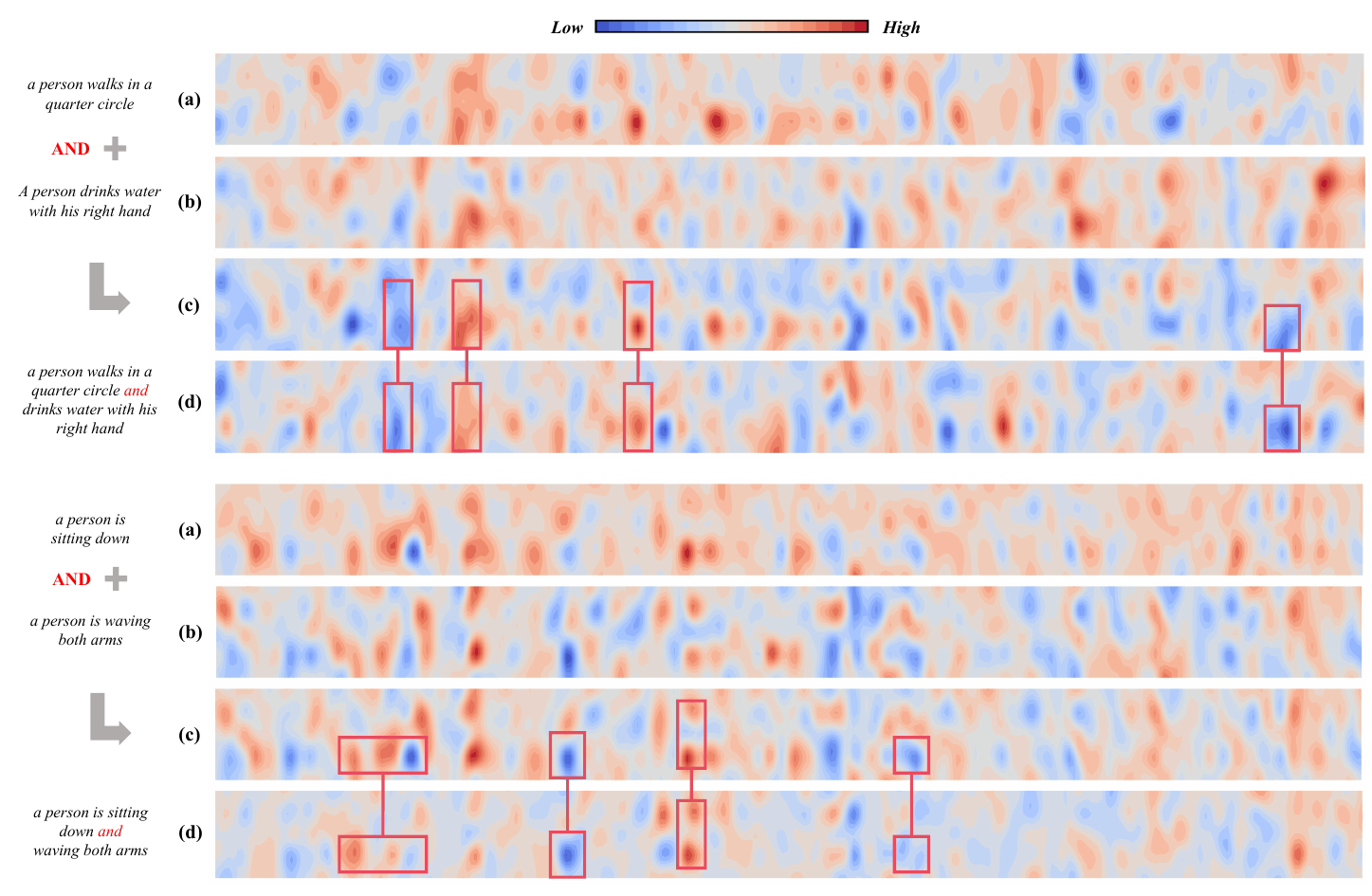}
\caption{\textbf{Visual results of energy distributions.} For a clear illustration, energy distributions are calculated with interpolation and Gaussian smoothing and then visualized as contour maps. (a) Concept 1, (b) Concept 2, (c) Compositional motion generation, (d) Multi-concept motion generation. Similar regions are highlighted in \textcolor{red}{red}.}
\label{fig:supp_energy}
\end{figure*}

We show additional contour maps of energy distributions in Figure~\ref{fig:supp_energy}. We provide two examples of the concept conjunction. We visualize the energy distributions of motion latent representations generated from the denoising autoencoder. Multi-concept generation combines concepts from both (a) and (b).
Compared with multi-concept generation (\ie, (d) in Figure~\ref{fig:supp_energy}), energy distributions of composed motion (\ie, (c) in Figure~\ref{fig:supp_energy}) show consistent high- and low-energy regions. This demonstrates that complex motion latent distributions can be composed of simple distributions, indicating that our method is explainable. Such results further explain the effectiveness of \textsc{EnergyMoGen}.

\section{More Details on Datasets and Evaluation Metrics}
\label{sec:supp_details_data_eval}
\subsection{Datasets}
Our experiments are conducted on three datasets: HumanML3D~\cite{guo2022generating}, KIT-ML~\cite{plappert2016kit}, and MTT~\cite{petrovich24stmc}.
\begin{itemize}
    \item \textbf{HumanML3D}~\cite{guo2022generating} is a large-scale text-to-motion generation benchmark that contains 14,616 human motions with 44,970 textual descriptions. The dataset is split with proportions of 80\%, 5\%, and 15\% for training, validation, and testing, respectively.
    \item \textbf{KIT-ML}~\cite{plappert2016kit} is another leading benchmark for motion generation from relatively short text. It has 3,911 finely annotated human motions, with 4888/300/830 for the training, validation, and test sets.
    \item \textbf{MTT}~\cite{petrovich24stmc} has 60 textual descriptions with body part annotations. The corresponding motions are collected from the AMASS dataset~\cite{mahmood2019amass}. Three texts are randomly composed based on body parts and motion duration through some conjunction words (\eg, ``and'', ``while''), resulting in a test set with 500 samples.
\end{itemize}
The three datasets use the same motion representation proposed in~\cite{guo2022generating}. 

\subsection{Evaluation Metrics}
We use evaluation models from Guo~\etal~\cite{guo2022generating} to measure the performance of text-driven human motion generation. We adopt the same metrics as previous works, including Frechet Inception Distance (FID) for motion quality, Retrieval Precision (R-Precision) and Multi-Modal Distance (MM-Dist) for text-motion consistency, and Diversity and MultiModality (MModality) for the diversity of generated motions. We denote two sets of features from the ground truth and generated motion as $m$ and $\hat{m}$, respectively.
\paragraph{FID.} The Fréchet Inception Distance (FID) measures the quality of generated motions by comparing their feature distributions to ground truth motions. It evaluates both the mean and covariance. Lower FID scores indicate better quality and closer resemblance to real data. FID can be calculated as:
\begin{equation}
\text{FID} = ||\mu_m - \mu_{\hat{m}}||^2 - \texttt{TR}(\Sigma_{m} + \Sigma_{\hat{m}} - 2(\Sigma_{m}\Sigma_{\hat{m}})^{\frac{1}{2}})
\label{formula:fid}
\end{equation}
where $\mu_{m}$ and $\mu_{\hat{m}}$ are mean of two sets of features. $\Sigma$ is the covariance matrix.

\paragraph{MM-Dist.}
MM-Dist is used to measure the distance between the generated motion and text directly:
\begin{equation}
\text{MM-Dist} = \frac{1}{N}\sum_{i=1}^{N}||m_i - \hat{m}_i||
\label{formula:mm-dis}
\end{equation}
where $N$ is the total number of motions.

\paragraph{Diversity.} To assess the diversity among motions generated by different textual descriptions in the test set, we randomly select 300 pairs of motions and compute this metric as follows:
\begin{equation}
\text{Diversity} = \frac{1}{300}\sum_{i=1}^{300}||\hat{m}_1 - \hat{m_2}||
\label{formula:diversity}
\end{equation}

\paragraph{MModality.} Similar to \textbf{Diversity}, MModality is used to measure the diversity among motions generated by the same text. We follow Guo~\etal~\cite{guo2022generating} to generate 30 motion samples from one text and randomly select two subsets, each containing 10 motions. The formulation of MModality is similar to the \textbf{Diversity} described above.

For compositional motion generation, we follow STMC~\cite{petrovich24stmc} to use R-Precision, TMR-Score, FID, and transition distance to evaluate the performance. Similar to MM-Dist, TMR-Score computes the cosine similarity between the generated motion embedding and text embedding with the TMR model~\cite{petrovich2023tmr}. TEACH~\cite{TEACH:3DV:2022} calculates the Euclidean distance between two consecutive frames as the transition distance.

\begin{figure}[t]
\centering
\includegraphics[width=\columnwidth]{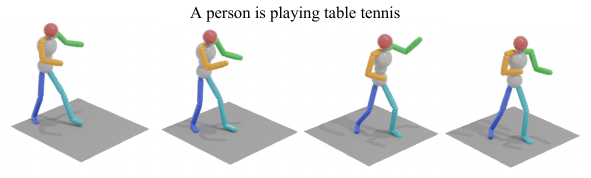}
\caption{\textbf{Failure case.}}
\label{fig:failure_case}
\end{figure} 

\section{Limitation and Failure Case}
\label{sec:limitations}
Existing latent diffusion models encode motion into a single (or a fixed number of) latent vector(s), which limits the use of per-frame composition algorithms.
We propose using an energy function to directly model the latent vector(s) encapsulating the overall features, \eg, temporal and skeletal features. 
The energy function (or energy) is additive. This property enables motion composition by composing energy functions (generated latent vectors) from different concepts together via conjunction and negation. However, our method still struggles with completely novel concepts. We provide a failure case in Figure~\ref{fig:failure_case}.

\end{document}